\documentclass[runningheads]{llncs}
\usepackage{graphicx}
\usepackage{comment}
\usepackage{amsmath,amssymb} %
\usepackage{color}
\usepackage{epsfig}
\usepackage{graphicx}
\usepackage{bm}
\usepackage{cite}
\usepackage{caption}
\usepackage{xcolor}
\usepackage{xspace}
\usepackage{wrapfig}
\usepackage{appendix}
\definecolor{citecolor}{HTML}{0071bc}
\usepackage[pagebackref=false,breaklinks=true,letterpaper=true,colorlinks,citecolor=citecolor,bookmarks=false]{hyperref}
\newcommand\norm[1]{\left\lVert#1\right\rVert}

\DeclareMathOperator*{\argmin}{arg\,min}

\def\eg{\emph{e.g.}\xspace}
\def\ie{\emph{i.e.}\xspace}

\begin{document}
\pagestyle{headings}
\mainmatter
\def\ECCVSubNumber{957}  %

\title{Deep Manifold Prior} %

\authorrunning{Gadelha, Wang, Maji} 
\author{Matheus Gadelha \quad Rui Wang \quad Subhransu Maji\\
{\tt\small \{mgadelha, ruiwang, smaji\}@cs.umass.edu}
}
\institute{University of Massachusetts, Amherst}

\maketitle

\begin{abstract}
  We present a prior for manifold structured data, such as surfaces of
  3D shapes, where deep neural
  networks 
  are adopted to reconstruct a target shape using gradient descent starting from
  a random initialization.
  We show that surfaces generated this way are smooth, with limiting behavior characterized by Gaussian processes, and we mathematically derive such properties for fully-connected as well as convolutional networks.
  We demonstrate our method in
  a variety of manifold reconstruction applications,
  such as point cloud denoising
  and interpolation, achieving considerably better results against competitive baselines while requiring no training data.
  We also show that when training data is available, our method allows developing
  alternate parametrizations of surfaces under the framework of
  AtlasNet~\cite{atlasnet}, leading to a compact network architecture
  and better reconstruction results on standard image to shape reconstruction
  benchmarks.
\end{abstract}

\section{Introduction}
\begin{wrapfigure}{r}{0.4\linewidth}
\vspace{-0.4in}
\centering
\includegraphics[width=1.0\linewidth]{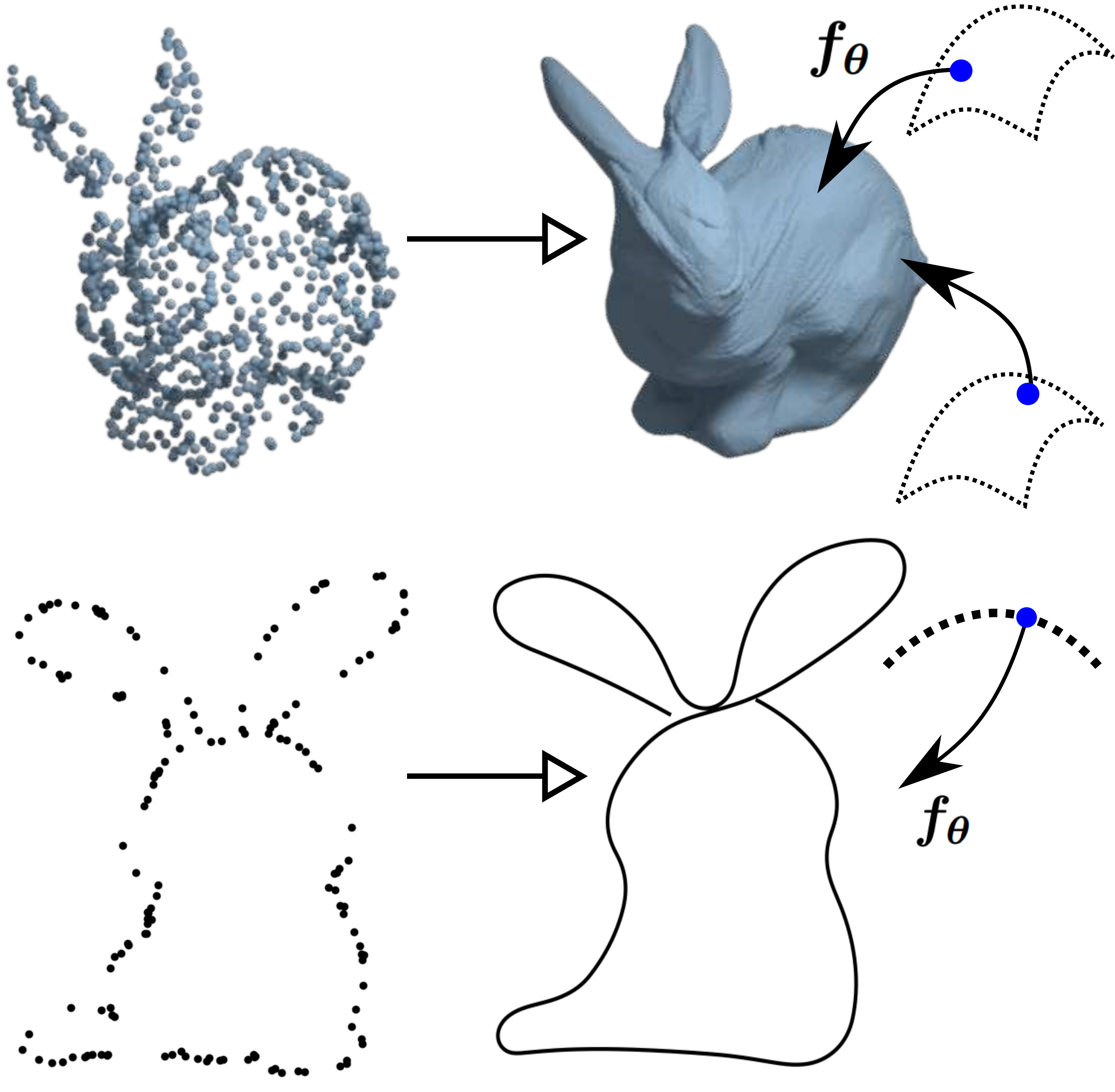}
\caption{\label{fig:visabstract}\small
\textbf{Deep manifold prior}. Points interpolated by using deep networks to map points in a 2D grid~(top) and 1D grid~(bottom) to the target shape (a 3D surface and a 2D curve respectively). The networks are randomly initialized and trained to minimize the Chamfer distance to the target.}
\vspace{-0.4in}
\end{wrapfigure}
In recent years a variety of approaches have been proposed to generate manifold data such
as surfaces of 3D shapes using deep networks.
The goal of this work is to characterize how the choice of the network architecture
impacts the properties of the resulting surfaces.
We present and analyze a \emph{deep manifold prior}, an approach to represent a
manifold as a collection of transformations (atlas) of an
Euclidean space parameterized using deep networks (Section~\ref{sec:method}).
We show that random networks induce smooth surfaces whose limiting
behavior can be understood in terms of a Gaussian process
(GP)~\cite{Neal,williams1997computing, cho2009kernel}.
We analyze how the different network architectures affect the
distribution of position, normals and
curvature of surfaces (Section~\ref{sec:gp}).
We also derive the properties of implicit surfaces 
induced by the level-set of a scalar field \{$f(x) = c$\} parameterized
using a deep network.

\begin{figure*}[ht]
\centering
\includegraphics[width=\linewidth]{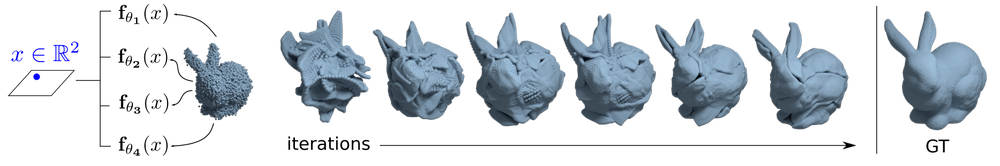}
    \caption{\label{fig:pipeline} \small \textbf{Manifold reconstruction pipeline.}
    Manifold parametrizations are encoded by neural networks ($\mathbf{f_{\theta_i}}$)
    and trained to minimize the reconstruction error with respect to the noisy target
    (left).
    Prior induced by the neural networks makes the generated surface much closer to
    the ground-truth (right), without ever seeing any additional training data.
}
\end{figure*}

As a concrete application we study the problem of 
interpolating and denoising point clouds sampled from contours or surfaces of
shapes, as seen in Figures~\ref{fig:visabstract} and~\ref{fig:pipeline}.
The manifold parametrization allows us to efficiently sample point
clouds, which can be combined with a Chamfer metric to measure a
reconstruction error with respect to the sampled data. 
We show that smooth surfaces are obtained when the parameters of the networks are learned to minimize the reconstruction error starting from a \emph{random initialization} (Figure~\ref{fig:pipeline}).
The approach is also effective for the level-set formulation, where the objective is to learn a deep network that correctly classifies points as \emph{inside} or \emph{outside} the surface.
However, an advantage of the explicit parametrization is that it does
not require the notion of what is inside.
In addition we introduce a \emph{regularization} that reduces self-intersections, overlaps, and distortion of the parametrization, which is desirable for applications such as texture mapping ~(Section~\ref{sec:method}).
Our approach requires \emph{no} prior learning, works across a range of 3D
shapes, and outperforms strong baselines for point cloud denoising, 
such as Screened Poisson Surface Reconstruction (SPSR) and Robust Implicit Moving Least Squares (RIMLS).
It is also more lightweight than approaches that
operate on volumetric representations of 3D shapes (Section~\ref{sec:experiments}).

Our analysis sheds lights on the impressive performance of several
recently proposed architectures for 3D surface generation, such as
MRTNet~\cite{mrt18}, AtlasNet~\cite{atlasnet}, FoldingNet~\cite{foldingnet}, and Pixel2Mesh~\cite{pixel2mesh}, as well as implicit surface approaches~\cite{chen2019learning,mescheder2019occupancy,genova2019learning,park2019deepsdf}.
These can be be interpreted
as different ways of parameterizing a manifold.
In particular, AtlasNet generates a 3D shape as a collection of surfaces, each represented as a
transformation of a unit grid using a fully-connected network.
However, the generated pieces exhibit significant overlap which
results in a poor surface reconstruction and is less desirable for
applying materials and textures to the surface (Section~\ref{sec:experiments}).
The proposed regularization alleviates this problem. 
Moreover, by replacing the fully-connected networks of AtlasNet with
convolutional variants we improve the performance on standard
benchmarks for shape generation~\cite{choy20163d} with networks that have a
fraction of the parameters, faster inference time, as well
as smaller memory footprint (Section~\ref{sec:experiments}).

\section{Related Work}
\paragraph{Manifold 3D shape generation}
3D shape generation is an active area of research with methods that
generate 3D shapes as volumetic representations such as occupancy grids~\cite{choy20163d, 3dgan, prgan, drcTulsiani17, tatarchenko2017octree, hie3dcnn, matryoshka}, 
signed distance functions~\cite{chen2019learning,mescheder2019occupancy,genova2019learning,park2019deepsdf}, mutliview depth and normals~\cite{LunGKMW17, Soltani17, tatarchenko2016multi, lin2018learning}, or point clouds~\cite{pcagan, fan2016point, latentpc, mrt18}.
Our work is closely related to techniques for generating 3D shapes
through a
predefined connectivity or parametrization structure over the surface of the shape. 
Pixel2Mesh~\cite{pixel2mesh} utilizes graph convolutional
networks to generate meshes that are homeomorphic to a sphere.
AtlasNet~\cite{atlasnet} and
FoldingNet~\cite{foldingnet} learn a parametrization of a
surface by adopting deep networks to transform
point coordinates in a 2D plane to the shape surface.
Specifically, each point is generated as $\big(f^1_{\theta}(x),f^2_{\theta}(x),f^3_{\theta}(x)\big)$
where $f^i_\theta$ is a deep network and $x=(x_1,x_2)$ is a point
in the unit grid.
Alternate approaches~\cite{chen2019learning,mescheder2019occupancy,genova2019learning,park2019deepsdf} represent the surface as the level-set of
a scalar field, $f(x) = 0, x\in \mathbb{R}^3$, e.g., of the
signed distance function.
While these have been applied for shape generation by training on 3D shape datasets, our goal is to analyze the role of these parameterizations as an \emph{implicit prior} for manifold denoising and interpolation tasks.

\paragraph{Deep implicit priors}
Our work is related to the deep
image prior~\cite{dip} that generates images as a
convolutional network transformation of a random signal on a unit
grid.
By optimizing the randomly initialized network to minimize a reconstruction loss with respect to the noisy target, their approach was shown to yield excellent denoising results.
Our approach generalizes this idea to manifold data, which is more appropriate for interpolating and denoising contours and surfaces  (see Figure~\ref{fig:dipcomp} for a comparison).
Our work is also related to the recently proposed deep geometric prior~\cite{williams2019deep}.
Their approach was used to estimate a surface from point cloud data by
partitioning the surface into small overlapping patches and reconstructing the local manifold
using a deep network.
Consistency in the overlapping regions was enforced by minimizing the Earth Movers distance (EMD). 
In contrast to their work, we learn a small collection of non-overlapping 
parametrizations (atlas) by minimizing a regularized term and Chamfer distance, which is much more efficient than EMD.
We also consider diverse tasks such as point cloud denoising, interpolation, and shape reconstruction across a category where the atlases needs to be consistent across instances. 
Finally, we present a theoretical analysis of the local properties of the
generated surface by analyzing its limiting behavior as a Gaussian
process.

\paragraph{Embedding a manifold}
Our work is related to techniques for embedding manifolds into a low-dimensional
Euclidean space (\eg, IsoMap~\cite{isomap} or LLE~\cite{lle}).
Our approach parameterizes the inverse mapping
from the Euclidean space to the data manifold using a deep network.
Interestingly, invertability can be guaranteed by using networks with easy to compute inverses (\eg,
NICE~\cite{dinh2014nice} or GLOW~\cite{kingma2018glow}). 
In computer graphics, a number of techniques have been developed for shape surface denoising and
reconstruction.
Screened Poisson Reconstruction~\cite{spsr} constructs an implicit surface
on a 3D volumetric grid based on oriented point samples by solving the Poisson equation.
Approaches based on Moving Least Squares~\cite{mls,
  alexa2003computing, rimls} reconstruct a surface by 
estimating an approximation of each local patch, similar to the deep geometric prior~\cite{williams2019deep} approach.
Our approach outperforms these baselines by a significant margin (Table~\ref{tab:denoising}).

\paragraph{Deep networks and Gaussian processes} A Gaussian process (GP) is
commonly viewed as a prior over functions. Let $T$ be an index set
(\eg., $T \in \mathbb{R}^d$), let $\mu(t)$ be a real-valued mean
function and $K(t,t')$ be a non-negative definite kernel or covariance
function on T. If $f \sim GP (\mu, K )$,
then, for any finite number of indices $t_1,...,t_n \in T$, the
vector $(f (t_i))^n_{i=1}$ is Gaussian distributed with mean vector $(\mu(t_i ))^n_{i=1}$ and covariance matrix $(K (t_i , t_j))^n_{i,j=1}$. 
 Neal~\cite{Neal} showed that a two-layer network with infinite number of
 hidden units approaches a GP.
 The mean and covariance of commonly used non-linearities have been
 derived in  several subsequent works~\cite{williams1997computing,cho2009kernel}. We use this
 machinery to analyze the limiting GP of deep manifold priors.

\section{Method}\label{sec:method}

\paragraph{Background} Our focus is to define priors over \emph{manifolds}.
We first introduce some basic notation.
A $n$-\emph{manifold} is a topological space $\mathcal{M}$ for which
every point in $\mathcal{M}$ has a neighborhood homeomorphic to the
Euclidean space $\mathbb{R}^n$.
Let $\mathcal{U} \subset \mathcal{M}$ and $\mathcal{V} \subset \mathbb{R}^n$ be open
sets.
A homeomorphism $\phi: \mathcal{U} \rightarrow \mathcal{V}, \phi(u) =
(x_1(u), x_2(u), ..., x_n(u))$ is a \emph{coordinate system} on
$\mathcal{U}$ and $x_1, x_2, ..., x_n$ are \emph{coordinate functions}.
The pair $\langle \mathcal{U}, \phi \rangle$ is a \emph{chart},
whereas $\zeta=\phi^{-1}$ is a \emph{parameterization} of $\mathcal{U}$.
An \emph{atlas} on $\mathcal{M}$ is a collection of charts
$\{\mathcal{U}_\alpha, \phi_\alpha\}$ whose union covers
$\mathcal{M}$. Intuitively, surfaces are 2-manifolds where as contours are 1-manifolds.
Thus the dimensionality of the input of the parameterization or the
output of the chart corresponds to the order $n$ of the manifold.
Atlases can be used to represent manifolds that cannot be decomposed
using a single parametrization (\eg, the surface of a sphere can be
diffeomorphically mapped to two planes but not one.)

\paragraph{General framework}
In our work we will replace the search over $\mathcal{U}$ by a search over
the parameters $\theta$ of the DNN $f_\theta$ that encodes the parameterization $f_\theta=\zeta=\phi^{-1}$.
More specifically, given a set of points $P \in \mathcal{M}$, we aim to recover the manifold $\mathcal{M}$ by computing the following:
\begin{equation}
    \theta^* = \argmin_\theta \mathcal{L}_C(f_{\theta, x \sim \mathbb{R}^n}(x), P).
\end{equation}
The approximated manifold can then be reconstructed in the domain on which it is embedded $f_{\theta^*}$.
In practice, we restrict $x$ to the unit hypercube $[0, 1]^n$.
Here $\mathcal{L}$ is a loss function that computes a discrepancy between sets.
Thus, reconstructing a manifold represented by an atlas of $k$ charts is done by computing the following: 
\begin{equation}
    \theta_1^*, \theta_2^*, ... \theta_k^* = \argmin_{\theta_1, \theta_2, ... \theta_k} \mathcal{L}_C(\bigcup\limits_{i=1}^{k} f_{\theta_i}(x), P)
\end{equation}

\paragraph{Parameterization}
We explore two choices of parameterizations of the coordinate function $f_\theta(x)$ as
a deep neural network.
The first uses a multi-layer perception (MLP) to represent the parameterization explicitly:
the network receives as an input a value $x \in \mathbb{R}^n$ and outputs the coordinates of point
in the manifold.
We use ReLU non-linearities throughout the network, except for the last layer where we
use $\tanh$.
This representation is analogous to the ones used in \cite{atlasnet, foldingnet}.
The second choice is to encode $\mathcal{M}$ directly through a
convolutional network $g(z)$, where $z$ is a stationary signal
(Gaussian noise).
We use 2D convolutional layers followed by ReLU
activations and bilinear upsampling, except for the last layer where
we use $\tanh$.
The convolutional parametrization induces a stationary prior (see Supplementary for details), and we observe the resulting architectures are more memory-efficient and compact than the first choice.

\paragraph{Loss function}
A key part of our method is computing a distance between two sets of points $P_1$ and $P_2$.
Such distance metric needs to be differentiable and reasonably efficient to compute, since the cardinality of the sets might be large.
Thus, similarly to previous work \cite{atlasnet, foldingnet, pixel2mesh, mrt18}, we employ the Chamfer distance $\mathcal{L}_C$ defined as follows:
\begin{equation}
\label{eq:chamfer}
    \mathcal{L}_C(P_1, P_2)= \!\sum_{p_1 \in P_1}\! \min_{p_2 \in P_2 }\norm{p_1 - p_2}_2^2 +\!
                   \sum_{p_2 \in P_2}\! \min_{p_1 \in P_1 }\norm{p_1 - p_2}_2^2. \nonumber
\end{equation}

\paragraph{Stretch regularization}
Representing the manifold as a set of multiple parameterizations output by
DNNs has some drawbacks.
First, there is no guarantee that the charts are invertible, which means
that a surface generated by $f_\theta$ might contain self-intersections.
Second, multiple charts might be representing the same region of the manifold.
In theory this is not a problem as long as overlapping regions are consistent. 
However, in practice this consistency is hard to achieve when point clouds are sparse and noisy.
We propose to alleviate those issues by penalizing the stretch of the computed parameterization.
Let $\mathcal{N}(w)$ be the neighborhood of $w$ in $\mathbb{R}^n$, the \emph{stretch regularization} $\mathcal{L}_S$ can be defined as follows:
\begin{equation}
    \label{eq:stretch}
    \mathcal{L}_S(\theta) = \mathbb{E}_{x\sim [0,1]^n}
        \left[\sum_{x^\prime \in \mathcal{N}(x)} \norm{f_\theta(x) - f_\theta(x^\prime)}_2^2 \right].
\end{equation}
Notice that we can compute the neighbors of $x$ ahead of time which makes
the computation significantly cheaper.
In practice, we sample $x$ from a set of predefined regularly spaced values in $[0,1]$ -- a regular grid in the 2D case.
Now we can define our full loss function as follows.
\begin{equation}
    \label{eq:objective}
    \mathcal{L}(\bm \theta) = 
    \mathcal{L}_C(\bm{f}_{\bm \theta, x \sim \mathbb{R}^n}(x), P) +
        \lambda \mathcal{L}_S(\bm \theta),
\end{equation}
where $\bm{\theta} = \theta_1, \theta_2, ... \theta_k$ and $\bm{f_\theta}(x) = \bigcup\limits_{i=1}^{k} f_{\theta_i}(x)$.

\paragraph{Manifolds as deep level-sets} An alternative approach is to represent $d$-manifold as the level-set of a scalar function over $d+1$ dimensions.
For example, a surface can be represented as the level set, $f(x) = 0$, where $x \in \mathbb{R}^3$. 
Prior work~\cite{chen2019learning,mescheder2019occupancy,genova2019learning,park2019deepsdf} has explored this approach to generate a 3D surface by approximating its signed distance function.
Level-set formulation can naturally handle shapes with different topologies, but require the knowledge of what is inside the surface, which can be challenging to estimate for imperfect point-cloud data.
In this work, we also characterize and experiment with the manifold prior induced
by the level-set of a deep network $f_\theta(x)=0$ initialized randomly.

\section{Limiting GP for the Deep Manifold Prior}\label{sec:gp}
\begin{figure*}[t]
\centering
\includegraphics[width=\linewidth]{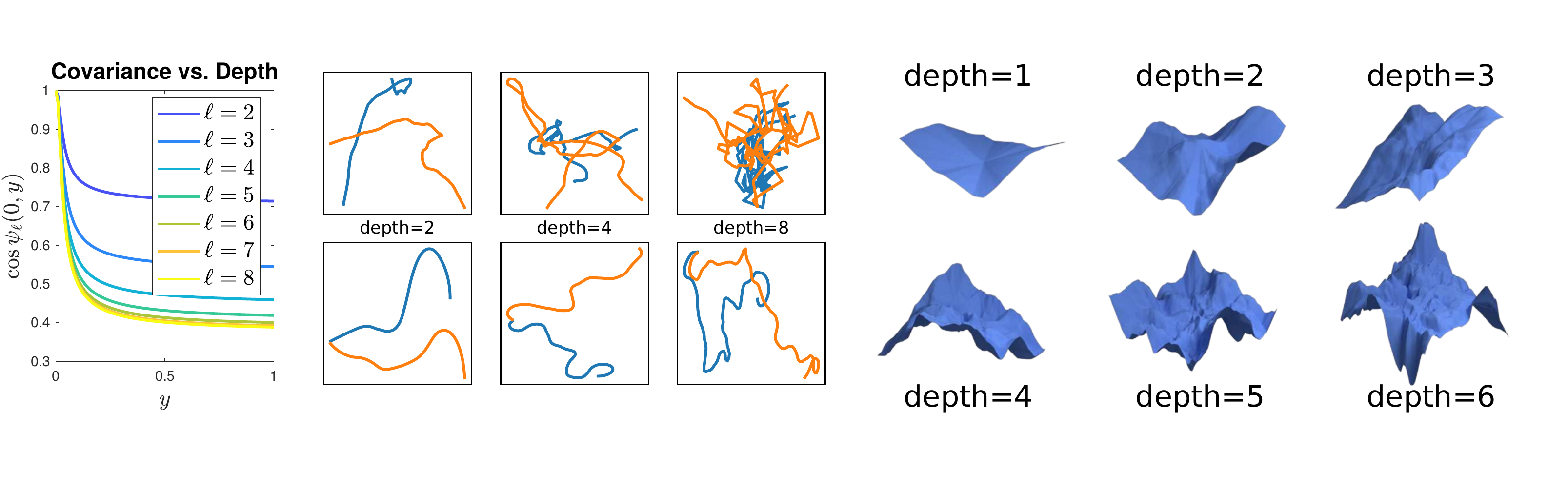}
\vspace{-30pt}
\caption{\label{fig:priorsamples} \small
  \textbf{Characterizing the deep manifold prior.}
  \textbf{(left)} a plot demonstrating the relationship between the network
  depth and the covariance function for the limiting GP. \textbf{(middle)}
  Random curves generated by the coordinate (top rows) and arc-length (bottom rows) parametrizations using deep networks with varying depths. \textbf{(right)} Random surfaces generated by deep networks of varying depths.} 
\end{figure*}

Consider the case when the manifold coordinates are parametrized using a deep network 
$f_\theta(x)$.
We show that random networks, \eg, whose 
parameters are drawn i.i.d. from a Gaussian distribution, produces smooth manifolds.
This is done by analyzing the limiting behavior of the function as a Gaussian
process.
In practice this is a good approximation to networks that are
relatively shallow and have hundreds of hidden units in each layer.

Concretely, the mean $\mathbb{E}_\theta[f_\theta(x)]$ and covariance
$\mathbb{E}_\theta[f_\theta(x) f_\theta(y)^T]$ of the parameterization 
characterize the structure of the generated manifold. 
For example, the covariance function of a smooth manifold decays
slowly as a function of distance in the input space compared to
a rough one.
Following prior work~\cite{Neal,williams1997computing,cho2009kernel},
we first derive the mean and covariance for a two layer network with a 
scalar output. We then generalize the analysis to vector outputs and multi-layer
networks.

Consider a two-layer fully-connected network on an input $x \in \mathbb{R}^n$. 
Let $H$ be the number of units in the hidden layer
represented using parameters $U = (u_1, u_2, \ldots u_H)$ where $u_j \in \mathbb{R}^n$
and the second layer has one output parameterized by weights $v \in
\mathbb{R}^H$. Denote the non-linearity applied to each unit as the
scalar function $h(\cdot)$. The output of the network is:  $f(x) = \sum_{k=1}^{H} v_k h(u_k^Tx).$
When the parameters $U$ and $v$ are
drawn from a Gaussian distributions $N(0, \sigma_u^2\mathbb{I})$ and
$N(0, \sigma_v^2\mathbb{I})$ respectively, we have:
\begin{equation*}
\mathbb{E}_{U,v}[f(x)] = \mathbb{E}_{U,v} \left[ \sum_{k=1}^{H} v_k
  h\left(u_k^Tx\right)\right] = 0, 
\end{equation*}
since $U$ and $v$ are independent and zero mean. 
Similarly, the covariance function $K(x,y)$ can be shown to be:
\begin{align*}
K(x,y) = \mathbb{E}_{U,v}[f(x)f(y)] = H\sigma_v^2 \mathbb{E}_{U} \left[ h\left(u_k^Tx\right)
  h\left(u_k^Ty\right) \right].
\end{align*}
This follows since each $u_k$ is drawn i.i.d, each
$v_k$ is independent and drawn identically from a Gaussian
distribution with zero mean.
The quantity $V(x, y)=E_u \left[ h(u^Tx)h(u^Ty)\right]$ can be
computed analytically for various transfer functions. 
Williams~\cite{williams1997computing} showed that when $h(t) = {\rm erf}(t) =
2/\sqrt{\pi}\int_{0}^{t}e^{-t^2}dt$, then 
\begin{align}
V_{{\rm erf}}(x, y) = \frac{2}{\pi}\sin^{-1} \frac{x^T\Sigma
  y}{\sqrt{\left(x^T\Sigma x \right)\left(y^T\Sigma y\right)}}. 
\end{align}
Here $\Sigma=\sigma^2 \mathbb{I}$ is the covariance of $u$.
For the ReLU non-linearity $h(t) = \max(0,t)$, Cho and Saul~\cite{cho2009kernel}
derived the expectation as:
\begin{align}
V_{\rm relu}(x, y) = \frac{1}{\pi}\Vert x \Vert \Vert y \Vert
\left(\sin \psi + (\pi -\psi)\cos \psi \right), 
\end{align}
where $\psi = \cos^{-1}\left(\frac{x^Ty}{\Vert x \Vert  \Vert y
  \Vert}\right)$. We refer the reader to~\cite{williams1997computing,cho2009kernel} for kernels
corresponding of other transfer functions.

An application of the Central Limit Theorem shows that by letting $\sigma_v^2$ scale as $1/H$ and $H \rightarrow \infty$,
the output of a two layer convolutional network converges to a
Gaussian distribution with zero mean and covariance 
\begin{align}
  K_1(x, y) = E_{U,v} \left[ f(x)f(y)\right] = V\left(x,y\right).
\end{align}

Hence the limiting behavior of the DNN can be approximated as a
Gaussian process with a zero mean and covariance function $K(x,y) =
V(x,y)$.

\paragraph{Extending to multiple outputs} The above analysis can be
extended to the case when the function $f(x)$ is vector
valued. For example a 2-manifold in 3D can be represented as $f(x) = (f^1(x), f^2(x), f^3(x))$, with $x \in \mathbb{R}^2$. 
In our case, the functions share a common backbone and each $f^i(x)$ is constructed from the outputs of the last hidden layer parameterized with
weights $v^i$, \ie, $f^i(x) = \sum_{k=1}^{H} v_k^i h(u_k^Tx).$
From the earlier analysis we have that each $f^i(x)$ has zero mean in
expectation. 
And the covariance between dimension $i$ and $j$ of $f$ is:
\begin{align*}
K_{1}^{i,j}(x,y) = \mathbb{E}_{U,v_i, v_j} \left[ f^i(x)f^j(y)\right] = V\left(x,y\right) \mathbf{1}[i=j].
\end{align*}
This follows from the fact that each $v_k^i$ is independent and drawn
from a zero mean distribution. 
Thus, the covariance is a diagonal matrix with entries $V(x,y)$ 
in its the diagonal.

\paragraph{Extending to multiple layers}
The analysis can be extended to multiple
layers by recursively applying the formula for the two-layer network.
Denote $K_\ell(x,y)$ as the covariance function of a scalar valued
fully-connected network with $\ell+1$ layers and $J(\theta) = \sin \theta
+ (\pi-\theta) \cos \theta$. Following~\cite{cho2009kernel} for the ReLU non-linearity we have the
following recursion:
\begin{align*}
K_{\ell+1}(x,y) = \frac{1}{\pi}\left(K_{\ell}(x,x) K_{\ell}(y,y)\right)^{1/2} J\left(\psi_{\ell}\right).
\end{align*}
Where $\psi_\ell(x,y) = \cos^{-1}
\frac{K_\ell(x,y)}{\sqrt{K_\ell(x,x)K_\ell(y,y)}}$ and $K_{0}(x,y) =
x^Ty$. Note that if in each layer we add a bias term sampled from a $N(0,
\sigma_b^2)$ the covariance changes to $K_\ell(x,y)
+ \sigma_b^2$ and the mean remains unchanged at zero. 

\subsection{Discussion and Analysis}
The above analysis shows that random networks induce certain priors
over the coordinates of the manifold. 
The effect of increasing the depth of the network can be seen by
visualizing how the covariance 
$\cos \psi_\ell (x, y)$ varies as a function of depth.
Figure~\ref{fig:priorsamples} plots $\cos \psi_\ell (x, y)$
at $x=0$ for a curve as a function of the depth of the network for
$\sigma_b=0.01$.
The covariance decays faster with depth, indicating
that the deeper networks produce manifolds with higher spatial frequencies (or curvatures).
This can also be seen in Figure~\ref{fig:priorsamples} which shows random curves (middle) from a
surfaces (right) for networks with varying depths. 

One potential drawback of fully-connected network parameterization is that the
generated manifold does not have a stationary (translationally invariant) covariance function. 
A covariance function $K(x,y)$ is stationary if it can be written as
$K(x,y) = k(x-y)$. 
On the other hand, a convolutional network that produces coordinates through a series of convolutional layers operating on
a random noise has a stationary covariance~\cite{cheng2019bayesian}. 
This is identical to the approach for generating natural images in the
deep image prior~\cite{dip} and we explore this alternative in Section~\ref{sec:exp_data}.

\paragraph{Normals and curvature} While we have shown that the outputs $f(x)$ induced by
random networks is a GP in the limit, what can be said about intrinsic properties
such as normals and curvature? 
Consider the curve $\gamma(t) = (x(t), y(t))$. 
Since derivatives are linear operators, it follows that 
distribution of derivatives, $\dot{x}$ and $\dot{y}$, are also Gaussian~\cite{solak2003derivative}.
The curvature is given by $\kappa = (\ddot{x}\dot{y} - \ddot{y}\dot{x})/(\dot{x}^2 + \dot{y}^2)^\frac{3}{2}.$
Unfortunately, since each of the derivatives converge to a zero mean
Gaussian distribution, the limiting distribution of the curvature $\kappa$ does
not exist. 
The pathology arises because the parameterization has a
speed ambiguity, \ie, replacing $t$
with any monotonic function of $t$ results in the same curve. 
To avoid this one can directly parametrize the derivatives as $\dot{x} =
\cos(f(t))$ and $\dot{y} = \sin(f(t))$ where $f$ is a deep network.
This is an arc-length (unit speed) parametrization since $\dot{x}^2 + \dot{y}^2 = 1$.
Once the derivatives are generated, the curve can be reconstructed by integration, \ie, $x = \int_0^t \cos(f(t)) dt$.
In this case the limiting distribution of the coordinates, normal, and
curvature all exist and are also GPs. We derive the mean and
covariance function in the Supplementary material.
Figure~\ref{fig:priorsamples}-middle shows draws from the GP with direct (top) and arc-length (bottom) parametrizations. 
One can see that arc-length parametrizations lead to more length-uniform curves.

Unlike curves, it is much more challenging to design arc-length parametrizations of surfaces. 
The difficulty arises due to the fact the gradients need to satisfy additional constraints for the surface to be
integrable~\cite{sussmann1973orbits}. 
Hence, we directly parameterized the coordinate function and proposed the stretch regularization to minimize distortion.
Alternatives ways of parameterizing the surface to satisfy properties such as conformality~\cite{nehari2012conformal} is left for future work.

\paragraph{Deep level-set prior} Finally, the GP analysis applies in a straightforward manner to the level-set formation $f_\theta(x) = 0$ where $f_\theta$ is a ReLU network 
mapping the 3D position $x \in \mathbb{R}^3$ to a scalar. The induced distribution over the
scalar field is a GP for random networks. Since for a differentiable function $f$ with non-zero gradient, the gradient is orthogonal to the level set, one can characterize the surface by analyzing the gradient field $\nabla f$. The limiting distribution over the gradient field is also a GP and one can estimate the mean and convariance functions by a similar analysis (see Supplementary material for details).
However, the training objective of the level-set prior is different from the explicit parameterization as the network must classify points as inside or outside the surface. This supervision can be challenging to obtain from noisy
data, especially for thin structures. We provide a comparison with this
approach in Section~\ref{sec:experiments}.

\section{Experiments}\label{sec:experiments}

\begin{table}[t]
\centering
\footnotesize
\tabcolsep=0.11cm
\begin{tabular}{|l|c|c|c||c|c|}
\hline
& Surface & Contour & Implicit & RIMLS\cite{rimls} & SPSR\cite{spsr} \\
\hline
bunny & \textbf{2.71E-04} & 6.64E-04 & 5.52E-04 & 1.43E-03 & 3.96E-04 \\
dragon & \textbf{4.18E-04} & 6.12E-04 & 1.20E-03 & 1.65E-03 & 1.46E-02 \\
car & \textbf{2.73E-04} & 4.57E-04 & 6.83E-02 & 1.50E-03 & 2.10E-03 \\
cup &  \textbf{2.59E-04} & 5.80E-04 & 2.64E-02 & 1.74E-03 & 1.00E-02 \\
mobius &  \textbf{3.51E-04} & 4.95E-04 & 3.26E-03 & 1.96E-03 & 1.89E-02 \\
chair &  \textbf{3.95E-04} & 4.22E-04 & 7.32E-03 & 2.09E-03 & 2.58E-02 \\
spiral &  1.05E-03 & \textbf{7.31E-04} & 1.64E-02 & 2.98E-03 & 7.90E-02 \\
ring &  5.69E-04 & \textbf{5.54E-04}& 4.81E-02 & 2.46E-03 & 3.76E-02 \\
\hline
avg. & \textbf{4.48E-04} & 5.65E-04& 2.13E-02 & 1.98E-03 & 2.36E-02 \\
\hline
\end{tabular}
\vspace{4pt}
\caption{ \small \label{tab:denoising} \textbf{Quantitative results for point cloud denoising.}
\emph{Surface}, \emph{Contour} and \emph{Implicit} represent different \emph{deep manifold priors} based on a 2-manifold, 1-manifold and level-set paramertization.
}
\vspace{-15pt}
\end{table}
\begin{table*}[t]
\centering
\scriptsize
\tabcolsep=0.11cm
\begin{tabular}{|l|c|c|c|c||c|c|}
\hline
& S1R & S8R & S1 & S8 & - & - \\
\hline
avg. & 4.48E-03 & \textbf{4.48E-04} & 2.75E-03 & 1.35E-03  & - & - \\
\hline \hline
& C1R & C8R & C1 & C8 & RIMLS\cite{rimls} & SPSR\cite{spsr} \\
\hline
avg. & 1.08E-03 & 5.77E-04 & 1.00E-03 & 5.82E-04 & 1.98E-03 & 2.36E-02 \\
\hline
\end{tabular}
\vspace{4pt}
\caption{ \small \label{tab:ablation} \textbf{Ablation studies.}
Comparison between different variations of our approach.
Naming follows the following convention: S corresponds to a 2-manifold parameterization (surface), whereas C corresponds to a 1-manifold (contour).
The following number (1 or 8) corresponds to the number of parameterizations.
A R letter is added if stretch regularization was used ($\lambda=1.0$).
}
\end{table*}

\begin{figure*}[t]
\centering
\includegraphics[width=\linewidth]{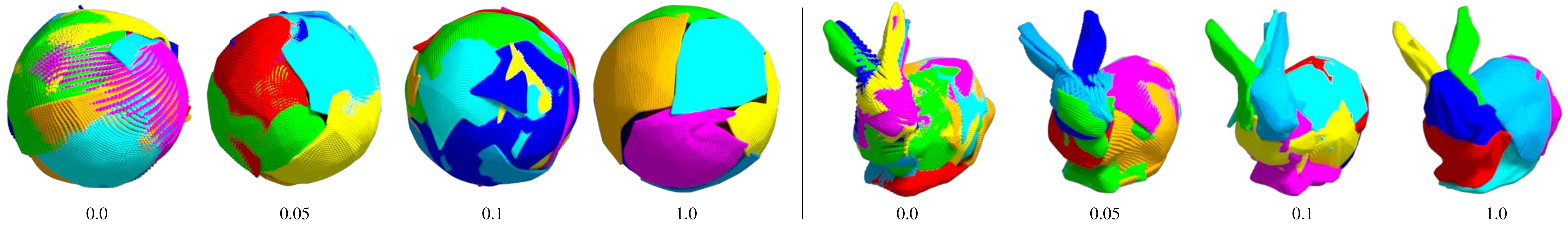}
	\caption{\label{fig:reg} \small
	\textbf{Effect of the regularization weight on the reconstructed manifold.} 
	For this experiment, we use our method to reconstruct a sphere using an atlas with 8 charts and render each one with a different color.
	Without any regularization, there is a significant amount of deformation applied to each surface (hence the space between the points) and a considerable amount of overlap between different parts. As the regularization weight increases, those aspects are
	noticeably reduced. 
	\small}
	\vspace{-12pt}
\end{figure*}

In this section we will present quantitative and qualitative results for
applying the manifold prior to multiple manifold reconstruction tasks.
All the experiments in this paper were implemented using Python 3.6 and PyTorch.
Computation was performed on TitanX GPUs. 

\subsection{Denoising and Interpolation}

\begin{figure*}[t]
\centering
\includegraphics[width=1.0\linewidth]{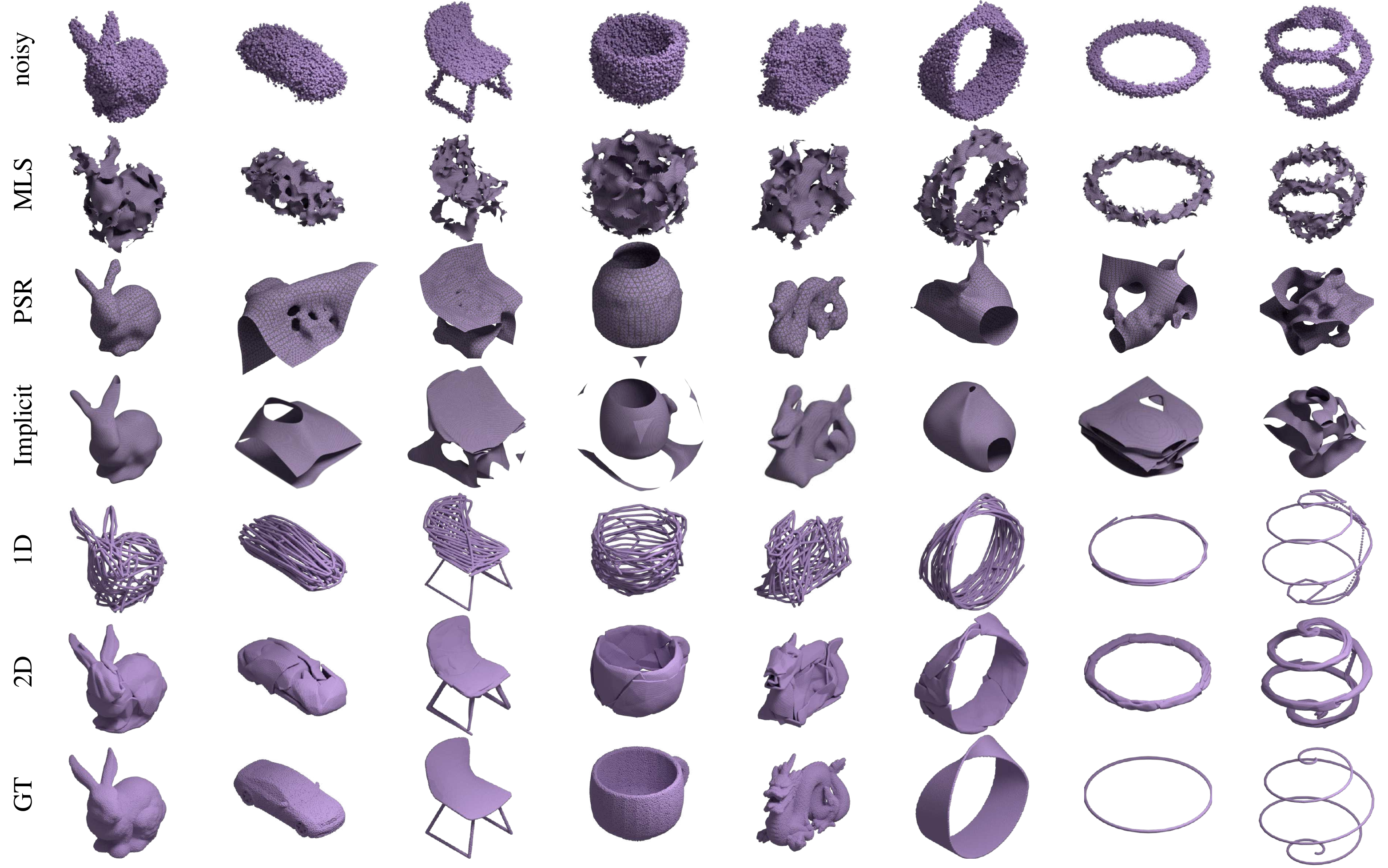}
	\caption{\label{fig:denoising} \small
	\textbf{Qualitative comparison between different denoising methods.}
	Rows display different methods, whereas columns display different shapes.
	Baseline methods do not perform as well as the deep manifold prior, even for closed surfaces like the bunny (first column) and the dragon (fifth column).
	As we can see, 2-manifold parameterizations are better for reconstructing surfaces, whereas 1-manifold counterparts reconstruct the curves (last two columns) more
	acurattely.
	\small}
\end{figure*}

\paragraph{Benchmark} Our benchmark consists of 8 different 3D shapes with diverse characteristics.
The shapes are normalized to fit a unit cube and 16K points are sampled on their surfaces.
The point positions are perturbed by a Gaussian noise with standard deviation $2\times10^{-3}$ and zero mean.
Figure~\ref{fig:denoising} shows the ground-truth shapes as well
as their noisy counterpart.
Since the level-set representation and the baseline methods 
(RIMLS~\cite{rimls},  SPSR~\cite{spsr}) 
require normal information, we estimate the normal for every point by using the local frame defined by its nearest neighbors.
We experimented multiple numbers of neighbors for both baselines and used the value that led to the best results: 20 neighbors for SPSR and the level-set representation, 30 neighbors for RIMLS.
The network used in the level-set representation follows the same architecture
and training protocol
as the one used for the explicit parametrizations (described in the next paragraph).
However, it is trained to predict every point as outside (+1) or inside the surface (-1).
Points with positive values are generated by translating every point in the point cloud along the normal direction for a distance $\epsilon=2\times10^{-3}$.
Points with negative values are generated in the same way, but applying a displacement
to the opposite direction.
For RIMLS, we used a relative spatial filter size of 10, 15 projection iterations and a volumetric grid with $200^3$ resolution.
For SPSR, we used an octree with depth 7 and 8 iterations.

\paragraph{Experimental setup}
Our method performs denoising by minimizing Equation~\ref{eq:objective}.
In this framework, $P$ is the noisy point cloud we are trying to reconstruct and $\bm{f_\theta}$ is a neural network.
In all experiments we use a neural network with 3 fully connected layers,
where the layers have 256, 128 and 64 hidden units, respectively.
The output of the networks is a point in $\mathbb{R}^3$.
The input can be either a point in $\mathbb{R}$ (1-manifold) or $\mathbb{R}^2$ (2-manifold).
We use $ReLU$ activations followed by batch normalization at each layer, except for the last, where we use a $tanh$ non-linearity.
We vary the architecture of $\bm{f_\theta}$ with respect to the number
of parameterizations (1 or 8) and dimensionality (1 or 2).
Additionally, we try each one of these architectural variations with
$\lambda=0$ and $\lambda=1.0$.
When using 8 parametrizations, 4096 points are sampled per parametrization.
When using just one parametrization, 16K points are sampled.
We optimize our objective through gradient descent using the Adam optimizer
with learning rate $10^{-3}$.
For evaluation, we uniformly sampled 16K points in the computed manifold (represented as a triangular mesh) and compute the Chamfer distance with respect to the ground-truth.

\paragraph{Results and discussion.}
Our methods significantly outperform the baselines for most of the shapes.
Quantitative results can be seen in Table~\ref{tab:denoising} and the qualitative results are shown in Figure~\ref{fig:denoising}.
The numbers are computed using 8 parametrizations (for surfaces and curves) and
$\lambda=1.0$.
A comparison between different variations of our approach is displayed in Table~\ref{tab:ablation}.
RIMLS, SPSR and level-set representations (\emph{Implicit} in Table~\ref{tab:denoising}) 
have trouble reconstructing point clouds with a significant amount of noise.
This is due to the fact that those methods rely on accurate surface normal estimates to infer
inside/outside regions of the shape.
Besides, RIMLS and methods based on implicit functions (SPSR and level-set representations) work better when dealing with closed surfaces. 
Shapes that are better approximated by contours (ring, spiral, chair's legs) are particularly challenging for those approaches.
On the other hand, the networks parametrizing explicit functions (\emph{Surface} and \emph{Contour} in Table~\ref{tab:denoising}) are able to adapt to different structures and
present a fair performance across a diverse set of shapes.

\begin{figure}
\centering
\includegraphics[width=0.6\linewidth]{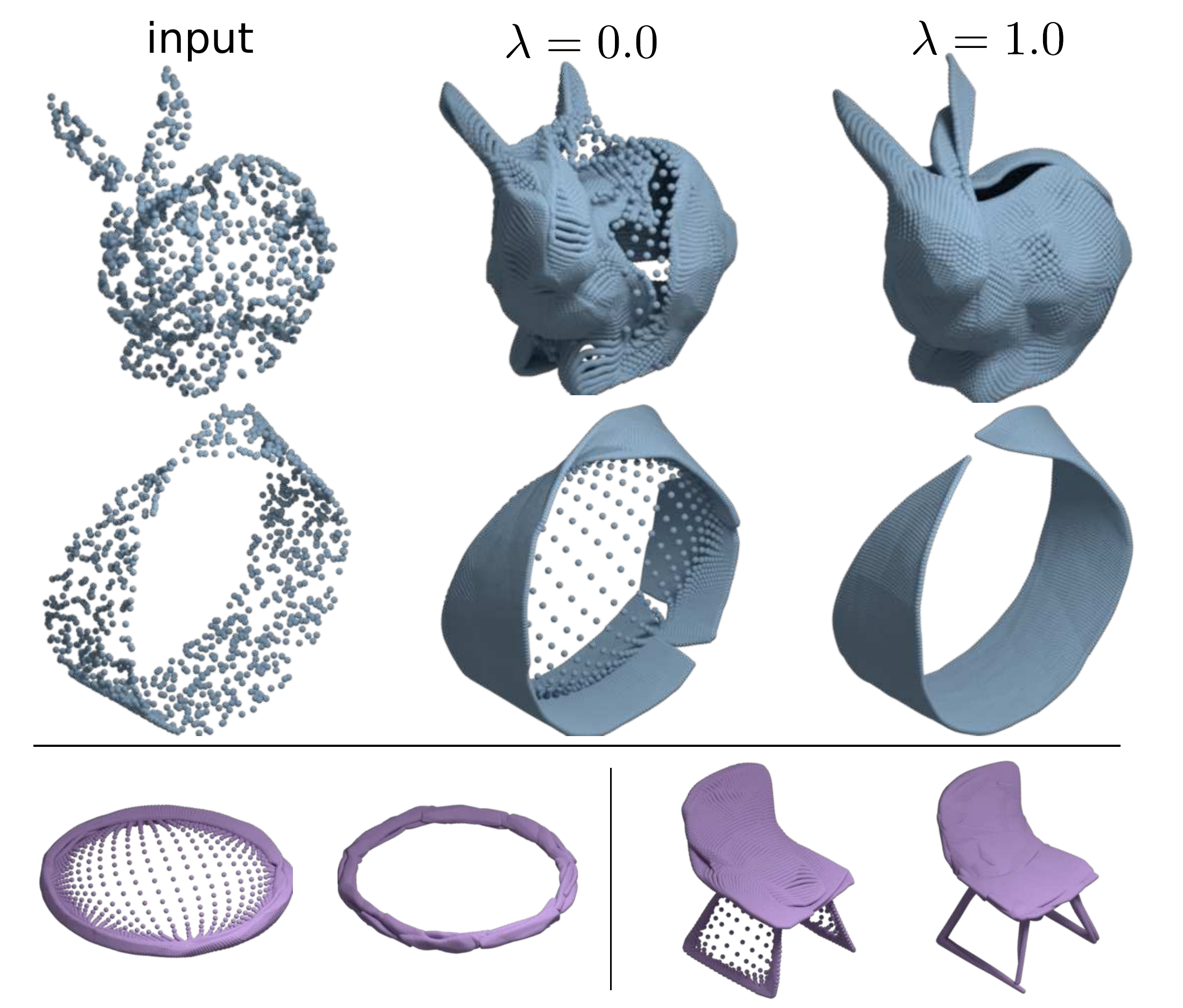}
    \caption{\label{fig:interp} \small 
    \emph{Interpolation results on the top.}
    Stretch regularization ($\lambda=1.0$) helps
    generate smoother surfaces.
    \emph{On the bottom, denoising using one vs. multiple parametrizations.}
   Shapes on the left were reconstructed using a single parameterization, 
    whereas shapes on the right used 8 parameterizations. 
    Using multiple parameterizations helps reconstruct complex shapes.
}
\vspace{-10pt}
\end{figure}

The results in Table~\ref{tab:ablation} suggest that using multiple parametrizations gives a better approximation than just using a single one.
This happens because complex shapes are easier to represent by multiple parametrizations.
For example, while using a single 2-manifold parametrization, the ring tends to be approximated by a disk, which significantly increases the reconstruction error when
the points are uniformly sampled over the final mesh.
This behavior is illustrated in Figure~\ref{fig:interp}.
Our ablation studies also indicate that using stretch regularization helps
parametrizations of both surfaces and contours.
Figure~\ref{fig:reg} shows the effect of stretch regularization for
two different shapes.
As the regularization weight increases, the overlap between different parameterizations becomes smaller.
When overlaps exist, the manifold representation is suboptimal --
the same regions are being generated multiple times.

\begin{figure}[t!]
\centering
\includegraphics[width=0.8\linewidth]{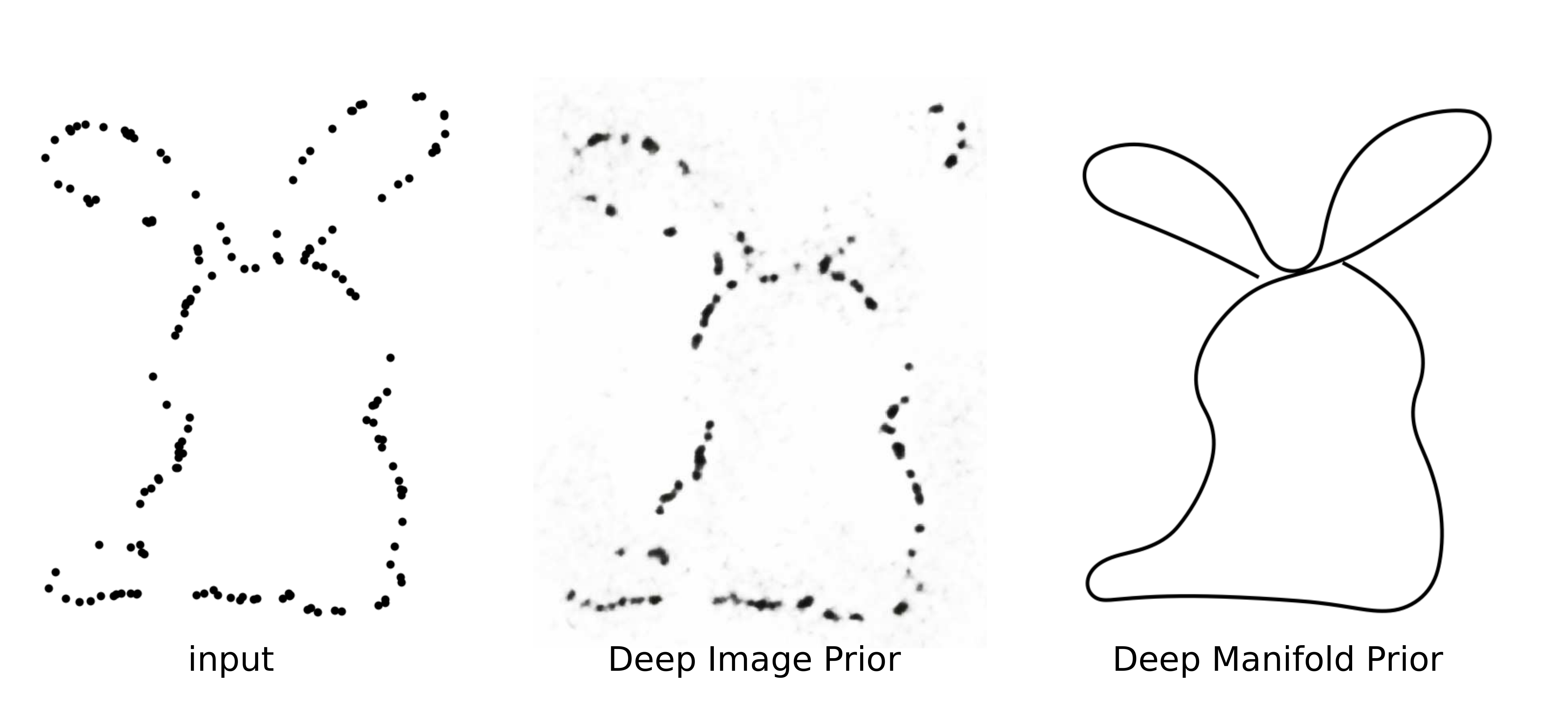}
	\caption{\label{fig:dipcomp} \small
	\textbf{Comparison to the deep image prior~\cite{dip}.}
	Image-based prior (middle) is not able to connect the dots in the input image (left).
	On the other hand, the manifold prior is able to reasonably interpolate the dotted
	drawing.
	}
	\vspace{-18pt}
\end{figure}

\paragraph{Interpolation} We also explored using the manifold prior for point cloud interpolation.
This experiment follows the same experimental setup as denoising.
However, instead of perturbing the points with Gaussian noise, we randomly select 1K points out of 16K.
Interpolation is performed by minimizing Equation~\ref{eq:objective}.
Results can be seen in Figure~\ref{fig:interp}.
For these experiments we use a single parameterization and include stretch regularization, without which the surface has holes and significant folds.
Our method is able to reconstruct reasonable surfaces from a small
set of points.

\paragraph{Comparison with the deep image prior}
We also compare our approach to the deep image prior~\cite{dip} for interpolating points in 2D
images.
Results are presented in Figure~\ref{fig:dipcomp}.
We use the same architecture from \cite{dip} while minimizing
the mean squared error with respect to the image pixels.
For the manifold prior, we use a single 1-manifold parameterization following
the architecture described before, differing only in the dimensionality of the output: points this this case are in $\mathbb{R}^2$ instead of $\mathbb{R}^3$.
Coordinates of the black pixels in the input image are used to form a point cloud and 
the manifold is computed by minimizing Chamfer distance with respect to it.

\begin{figure}
\centering
\includegraphics[width=0.8\linewidth]{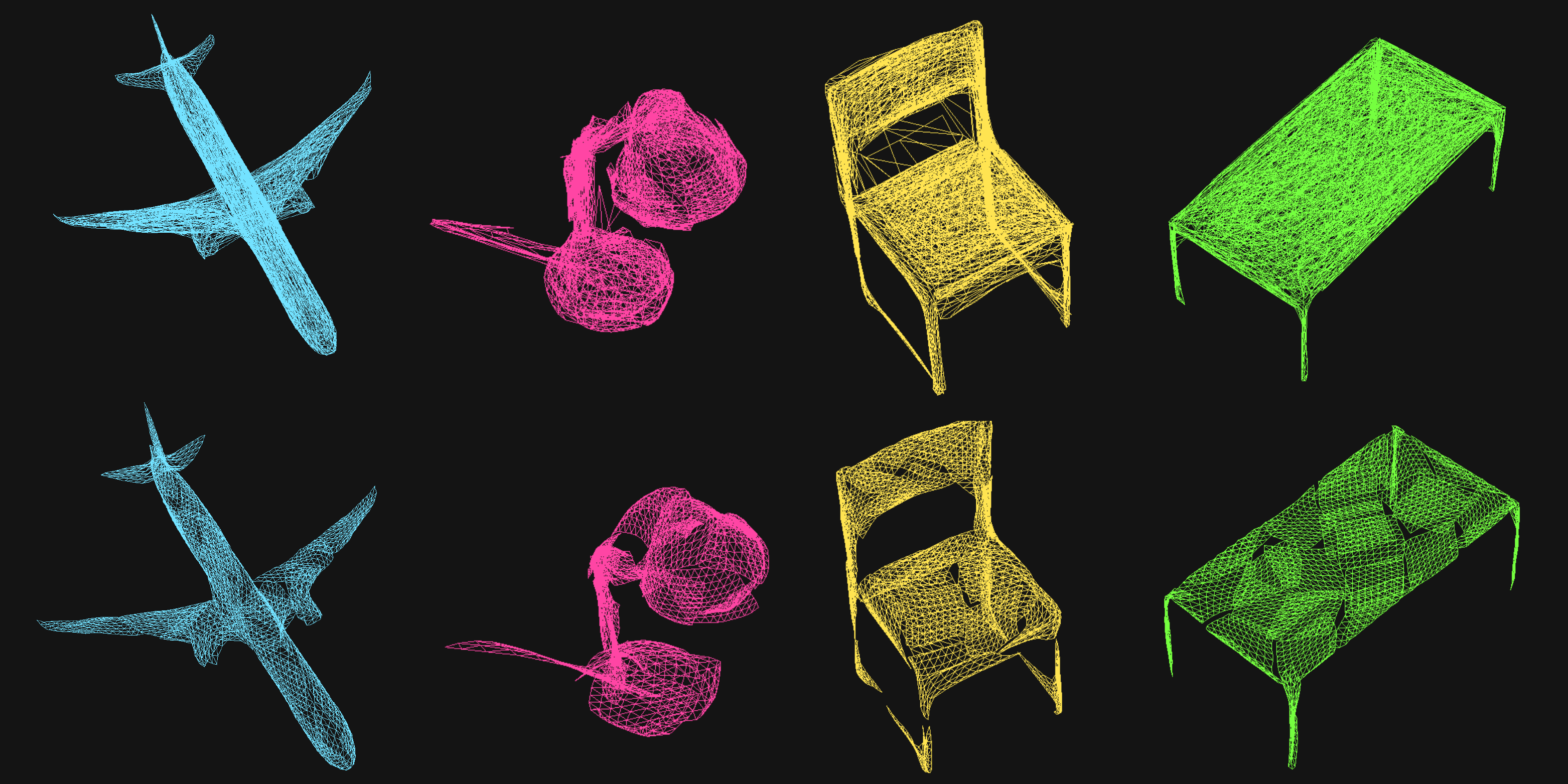}
    \caption{\label{fig:aestretch} \small \textbf{Autoencoder results}.
    Results on using AtlasNet~\cite{atlasnet} trained w/o (top) and w/ (bottom) stretch regularization.
    The latter results in meshes with reduced deformation and overlap, and removes artifacts where the chair's back is incorrectly filled.
}
\end{figure}

\subsection{Learning from data}\label{sec:exp_data}
Finally, we show how the insights presented in the earlier sections, in particular convolutional parameterization and stretch regularization, can also improve generative models of 3D shapes when trained on a large collection of shapes. 

To measure the effect of the stretch regularization in a learning-based scenario,
we train a model using the same architecture as AtlasNet~\cite{atlasnet} on a subset of $50,000$ shapes across $13$ categories of the ShapeNet dataset~\cite{chang2015shapenet}.
Adding stretch regularization did not significantly impact the Chamfer metric -- error of $1.46 \times 10^{-3}$ and $1.47 \times 10^{-3}$ with and without regularization.
However, the results are qualitatively better. 
As seen in Figure~\ref{fig:aestretch} the regularization reduces the stretch and overlap of the generated surfaces, and eliminates artifacts where holes are incorrectly filled.

We also train a convolutional decoder with stretch regularization on the single-view reconstruction benchmark~\cite{choy20163d}.
Our approach called ConvAtlas is compared against AtlasNet and MRTNet~\cite{mrt18} in Table~\ref{tab:svr}.
For a fair comparison, we use 4K points for evaluation across all methods.
ConvAtlas outperforms both approaches in terms of per-category and per-instance error, and also leads to more compact models.
Per-category results and experimental details are in the Supplementary material.

\begin{table}[t!]
\tabcolsep=0.11cm
\centering
\small
\begin{tabular}{|l|c|c|c|}
    \hline
    Architecture &  mean/cat. & mean/inst. & \#params. \\
    \hline
    MRTNet & 4.80 & 4.26 & 81.6M\\
    AtlasNet & 4.74 & 4.38 & 42.6M\\
    ConvAtlas & \textbf{4.53} & \textbf{4.00} & \textbf{ 14.5M}\\
    \hline
\end{tabular}
\vspace{4pt}
\caption{\small \textbf{Quantitative results for single-view image-to-shape reconstruction.} The table reports the mean Chamfer distance metric (scaled by $10^3$) computed per category and per instance.
  }
  \label{tab:svr}
  \vspace{-12pt}
\end{table}

\vspace{-3pt}
\section{Conclusion}
\vspace{-4pt}
We presented a manifold prior induced by deep neural networks.
Our experiments show that the prior can be effectively used for a variety of manifold reconstruction
tasks: denoising, interpolation and single-view reconstruction.
Besides, we analyzed the influence of the architecture in the characteristics of the prior
by posing the models as GP.
In conjunction to the prior induced by deep networks, we showed that using a stretch regularization procedure enables
better manifold approximation and improves the quality of the generated meshes, reducing large deformations and overlaps
between different parameterizations.

\vspace{6pt}\noindent\textbf{Acknowledgements.}
The authors would like to thank Daniel Sheldon for helpful discussions
related to Gaussian processes.
This work is supported in part by  
NSF grants \#1908669 and \#1749833. Our experiments were performed in the
UMass GPU cluster obtained under the Collaborative Fund managed by the
Massachusetts Technology Collaborative.

\appendix
\section{Convolutional Parametrizations}

In the main paper, we experimented with fully connected architectures
for representing manifold parametrizations.
However, parametrizations represented by convolutional architectures
also induce a prior useful for manifold reconstruction tasks.
In this section, we show experiments with denoising and single-view
reconstruction.
We start by defining a \texttt{ConvBlock}, which consists of a bilinear upsampling layer followed by a 2D-conv, batch normalization~\cite{batchnorm} and Leaky ReLU activation (slope=$0.2$).
Every convolutional layer uses filter size $3 \times 3$, stride $1$ and the number of filters is exactly half the number of its input channels.
In other words, at every \texttt{ConvBlock}, the output tensor spatially doubles the size of its input tensor, but only has half the number of channels.
This pattern follows throughout the whole network, except for the last layer, where the output layer always have 3 channels, representing the $(x,y,z)$ point coordinates.

\subsection{Denoising}

\begin{figure}[ht!]
    \centering
    \vspace{-10pt}
    \includegraphics[width=1.0\linewidth]{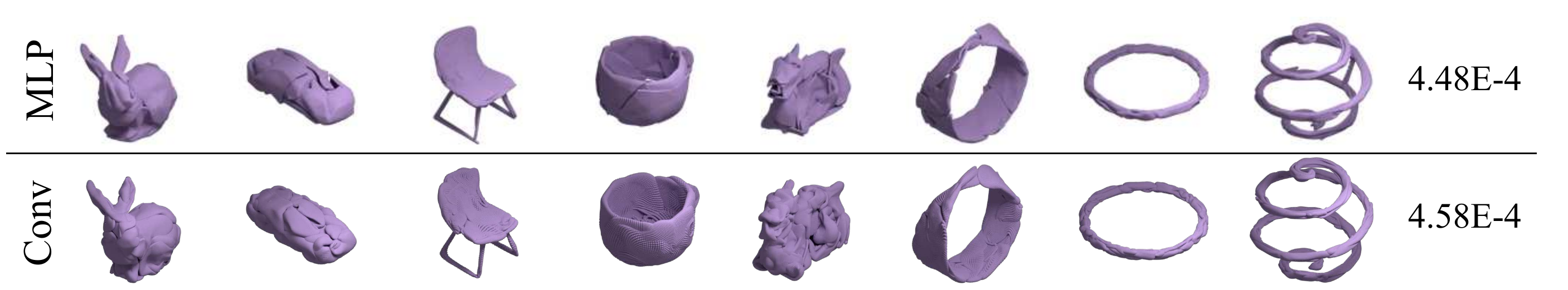}
    \vspace{-20pt}
    \caption{\label{fig:rebuttal}\small
    Comparison of Conv and MLP networks for denoising.
    Average error across shapes to the right. Both models use 8
    parametrizations and stretch regularization. Zoom for details.
    }
\end{figure}

The denoising experiments follow the same procedure described in the
main paper, except for the network architecture.
Instead of using a fully connected model, we employ a network with 3 \texttt{ConvBlock}s,
starting from an input tensor with shape $4\times4\times512$ whose values are drawn from
a standard gaussian distribution.
The output of each parametrization is a tensor with shape $32\times32\times3$, which we can treat
as a point cloud with 1024 and use Chamfer distance in the same way as described in Section 5.
We also use the position of the points in the output tensor to define the local neighborhood
utilized in the stretch regularization.
Results are presented in Figure~\ref{fig:rebuttal}.
As we can see, convolutional parametrizations also induce a useful prior for manifold
reconstructions and, similarly to the other parametrizations, it is significantly
better than the baselines.
Quantitatively, using convolutional parametrizations in the denoising yields slightly
worse results than using fully connected networks -- in terms of Chamfer distance, 4.58$\times10^{-4}$ vs. 4.48$\times10^{-4}$.

\subsection{Single-view Reconstruction}

\begin{figure*}[t]
\centering
\includegraphics[width=0.9\linewidth]{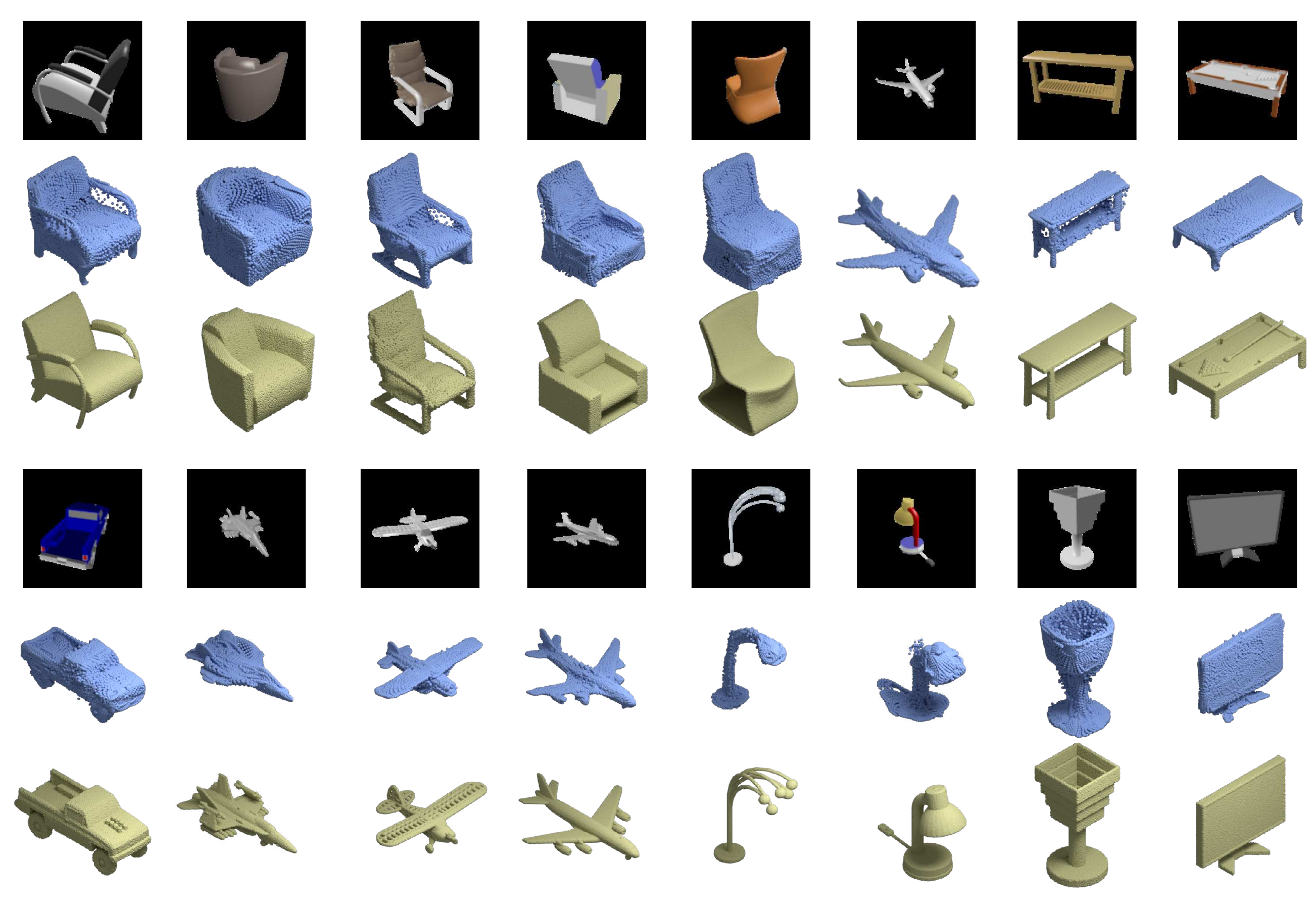}
\vspace{-8pt}
	\caption{\label{fig:data} \small
	\textbf{Image-to-shape reconstruction results from the test set.} The images shown are the input (black background), our results (32K points, rendered blue), and ground truth (rendered in light green). Qualitatively, our method is able to generate high-resolution point clouds faithfully capturing fine geometric details such as the chair legs, arms, airplane engines, monitor stands etc.}
	
\end{figure*}

In this subsection we present quantitative and qualitative results for single-view image-to-shape using convolutional paramterizations.
We also train a convolutional decoder with stretch regularization on the single-view reconstruction benchmark~\cite{choy20163d}.
This follows the same experimental setup as previous papers \cite{fan2016point, choy20163d, atlasnet, mrt18}.
However, unlike AtlasNet~\cite{atlasnet}, our network is trained in one stage, without the need to
train the decoder in an auto-encoder setting before fine-tuning it with an image CNN in a second step.
We used Adam optimizer~\cite{kingma2014adam} with learning rate of $10^{-3}$. 
The model is trained for 40 epochs and the learning rate is divided by 2 every 5 epochs.
We use ResNet-18 as image encoder and 32 convolutional parameterizations.
Even though we use more parameterizations than AtlasNet, the total number of parameters is smaller (see Table~\ref{tab:param}.
The evaluation results per category are presented in Table~\ref{tab:svr}.
Compared to MRTNet, our model performs better in 12 out of 13 categories. Compared to AtlasNet, our method is better or ties (the firearm category) in 7 out of 13 categories.
Overall our approach outperforms AtlasNet in per-category mean by 0.21, a relative improvement of 4.4\%. Also note that our model outperforms AtlasNet mainly in categories with
a large number of examples (tables, cars, airplanes, chairs). As a result, if average over instances, our method has a per-instance mean of 4.0, vs. 4.38 by AtlasNet -- a relatively improvement of  8.7\%.

\begin{table*}[h!]
\centering
{
\small
  \begin{tabular}{l|c|c|c|c|c|c|c|c|c|c|c|c|c||c}
   &  pla. &  ben. &  cab. &  car &  cha. &  mon. &  lam. &  spe. &  fir. &  cou. &  tab. &  cel. &  wat. &  \textbf{mean} \\
  \hline
 AtlasNet~\cite{atlasnet} &
2.17 &
\textbf{3.39} &
\textbf{3.93} &
3.40 &
4.56 &
\textbf{5.05} &
12.24 &
8.79 &
\textbf{2.15} &
\textbf{4.58} &
4.15 &
\textbf{3.25} &
3.93 &
4.74\\
 
 MRTNet~\cite{mrt18} &
2.25 &
3.68 &
4.73 &
\textbf{2.55} &
4.06 &
6.07 &
11.15 &
8.84 &
2.25 &
4.98 &
4.45 &
3.72 &
3.64 &
4.80\\

Ours (32 dec.)&
\textbf{2.06} &
3.40 &
4.46 &
2.60 &
\textbf{3.76} &
5.94 &
\textbf{10.66} &
\textbf{8.38} &
\textbf{2.15} &
4.64 &
\textbf{3.96} &
3.45 &
\textbf{3.40} &
\textbf{4.53} \\
  \end{tabular}
  }
\caption{\small \textbf{Quantitative results for single-view image-to-shape reconstruction.} The table reports Chamfer distance metric (scaled by $10^3$) computed per category, and the mean of all categories.
For each method 4K points were used to compute the distance. %
  }
  \label{tab:svr}
\end{table*}

\paragraph{Ablation studies.} Table~\ref{tab:alt} shows a quantitative comparison between a few architectural variations.
We start by analyzing a variation of our network that generates the same number of points (using a single decoder) as MRTNet (4K points) and the same image encoder (vgg-16).
The performance of this variation is 0.05 worse than MRTNet, but it has an order of magnitude less parameters than MRTNet.
Another variation is to still use a single decoder but generate a higher-resolution point cloud (16K points).
This variation results in improved Chamfer distance, by 0.1, than the first variation, indicating that the increased resolution does improve reconstruction accuracy. Again, even when the number of generated points is higher than 4K, our evaluation is done by randomly selecting 4K points, for fair comparison. The last row in the table is our default setting (32 decoders outputting a total of 32K points). 
The number of network parameters are reported in Table~\ref{tab:param}. Even though the number of points our network generates is 8 times that of MRTNet, its size is only about 1/6 of MRTNet, since our network does not need to represent multiple resolutions at each layer. Compared to AtlasNet, our network is about 1/3 of its size, due to the efficiency of using a fully convolutional architecture. Despite using a much smaller number of parameters, our network outperforms MRTNet (in terms of Chamfer distance metric) by 0.27, and AtlasNet by 0.21.

\begin{table}
\centering
{
\begin{tabular}{|l|c|c|}
    \hline
    Architecture &  mean/cat. & mean/inst. \\
    \hline
    MRTNet & 4.80 & 4.26 \\
    1 dec./vgg16/4k & 4.85 & 4.30\\
    1 dec./res18/16k & 4.75 & 4.22\\
    32 dec./res18/32k & \textbf{4.53} & \textbf{4.00}\\
    \hline
\end{tabular}
}
\caption{\label{tab:alt} \small
    \textbf{Architecture variations and evaluation results.} The table reports per-category mean and per-instance mean for MRTNet, and three variations of our methods: single decoder with 4K output points, 16K output points, and 32 decoders with 32 output points. For all cases, the Chamfer distance is calculated using 4K sample points, and results are scaled by $10^3$.
}
\end{table}

\begin{table}
\centering
{
\begin{tabular}{|l|c|}
    \hline
    Method &  \#parameters \\
    \hline
    AtlasNet & 42.6M \\
    MRTNet & 81.6M \\
    \hline 
    Ours (1 dec.) & 2.49M \\
    Ours (1 residual dec.) & 5.79M \\
    Ours (32 dec.) & 14.5M  \\
    \hline
\end{tabular}
}
\caption{\label{tab:param} \small
    \textbf{Comparing the \# of network parameters.}
}
\end{table}

\paragraph{Qualitative Results.} Figure~\ref{fig:data} shows image-to-shape reconstruction results for images from the test dataset. Overall our method is able to accurately capture fine geometric details such as the chair legs, arms, airplane engines, monitor stands etc. The number of points (32K) is considerably higher than previous work (e.g. 1K by~\cite{fan2016point} and 4K by~\cite{mrt18}). Some specific shapes, such as lamps and jet fighters, present significant challenges for the network as the input images do not contain all the visual details. Nonetheless our method is able to produce a reasonable approximation.

\begin{figure*}[t]
\centering
\includegraphics[width=0.9\linewidth]{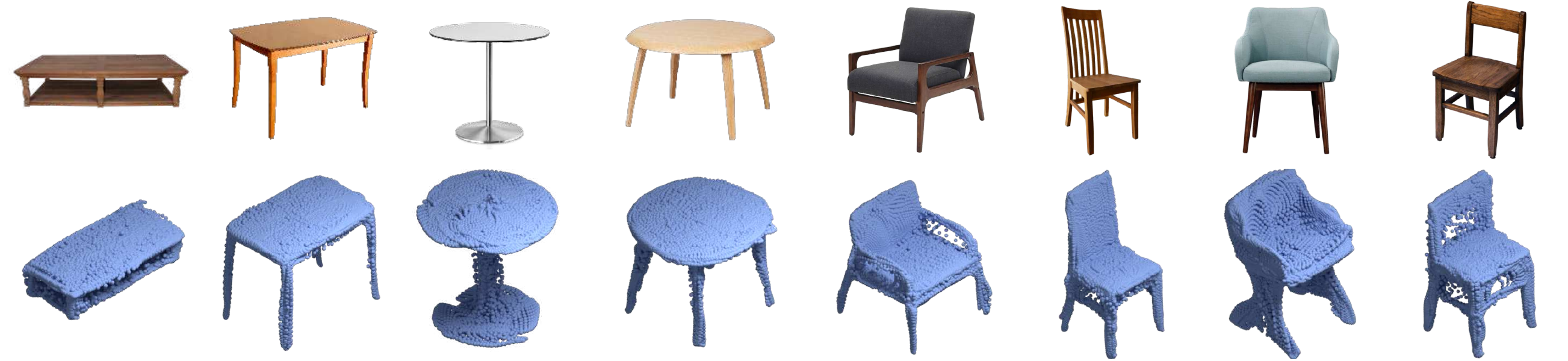}
\includegraphics[width=0.9\linewidth]{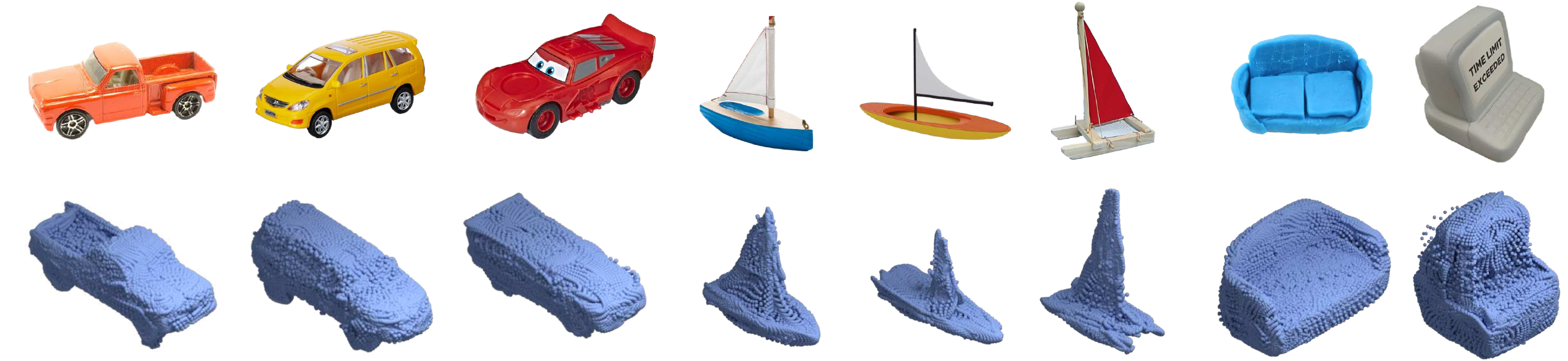}
	\caption{\label{fig:toy} \small
	\textbf{Image-to-shape reconstruction results on Internet photos.} We test our method on real photos downloaded from the Internet and the results are rendered in blue. 
    The test images here are considerably different from the training set. Our method achieves reasonable results with accurate geometric details. The last image (computer) represents a category that has not been seen during training.
    }
\end{figure*}

\paragraph{Test on real images.} The test set images are synthetically rendered and as such they look similar to the training images. To evaluate our method on real images we use photos downloaded from the Internet, as shown in~\cite{mrt18}. They are processed by removing the background so only the foreground object remains. Figure~\ref{fig:toy} shows the results. The top row in the figure shows furniture objects, which demonstrate that even though the network is trained using synthetic images rendered with artificial lighting and materials, the model is able to generalize well to real shading, lighting, and materials. The second row shows additional objects where the shading is considerably different from training images. In particular, the last image (desktop computer) is in a category that the training has never seen. Nonetheless the reconstructed shape is reasonable.

\begin{figure*}[h!]
\centering
\includegraphics[width=0.8\linewidth]{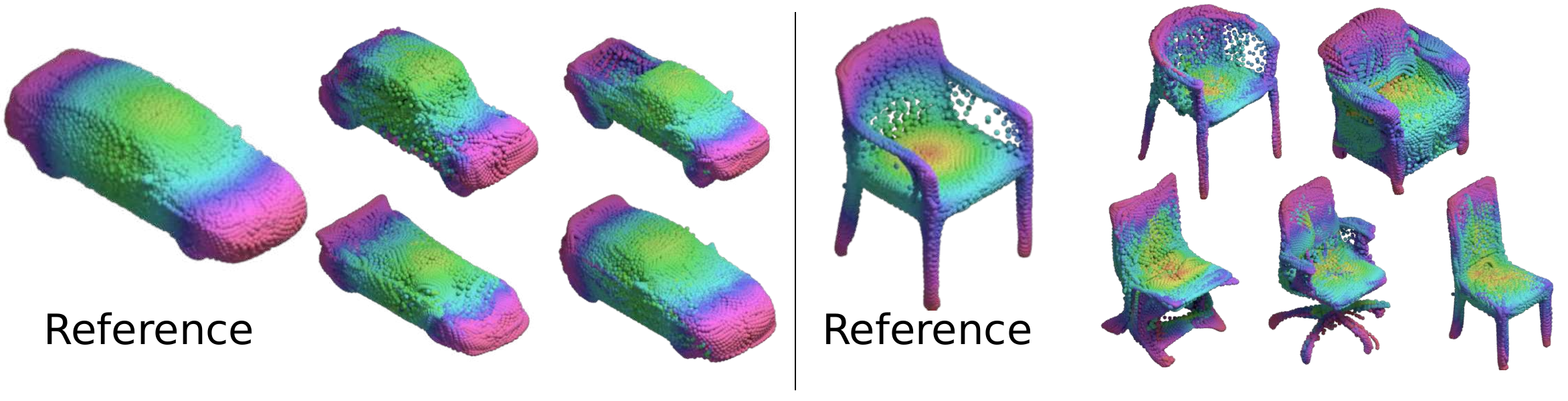}
	\caption{\label{fig:corresp} \small
	\textbf{Visualizing Shape Correspondences.}
	Our network learns approximate shape correspondences even though the training is not supervised with such information. The shapes shown here are generated by 32 decoders.
	}
\end{figure*}

\paragraph{Shape correspondence.} Once trained, our network learns to generate shapes with corresponding structures.
We demonstrate this with the following experiment.
First, we randomly select a point cloud generated by our network and call it a reference shape.
Then, we assign every point in the reference shape a color, where the hue is computed based on the point's distance to the center of gravity of the object.
Then this color assignment is propagated to the other point clouds, such that a point at index $(i,j)$ in the output tensor is assigned the same color as the point on the reference shape at the same index.
The resulting colorized point clouds are shown in Figure~\ref{fig:corresp}. Similar color indicates similar index range in the output tensor.
Note that even though the network is not trained explicitly with point correspondences as
supervisory signal, it learns to generate corresponding parts in the same
regions of the output tensor, as can be seen around
the tips of the chairs' arms,  legs and  seats.

\section{Limiting distribution for the curvature}

We start by parameterizing the derivative of a space curve $\dot{x} = \cos(f(t))$ 
and $\dot{y} = \sin(f(t))$ where $f$ is a neural network. 
From the standard analysis we know that $f(t)$ converges to a Gaussian with mean $\mu$ and kernel $k(\cdot,\cdot)$. 
Without loss of generality we can assume that the mean $\mu$ is such that $\cos(\mu) \neq 0$ and $\sin(\mu)\neq 0$. 
This can be achieved by adding a fixed bias term $\mu$ to the output of the last layer. 
To compute the limiting distribution of $\ddot{x}$ and $\dot{y}$ we apply the first order delta method to obtain:

\begin{align}
\dot{x} &\rightarrow {\cal N} (\cos(\mu), \sigma^2\sin^2(\mu)), \\
\dot{y} &\rightarrow {\cal N} (\sin(\mu), \sigma^2\cos^2(\mu)). 
\end{align}

Note we can only apply the first order delta method when the derivatives are not zero. 
Hence we assumed that $\mu$ is set to be a quantity which has this property. 
Otherwise we need the second-order delta method and the resulting distribution would be $\chi^2$ for one of the derivatives.

Since the derivative is a linear operator it follows that $\ddot{x}$ and $\ddot{y}$ are also GPs. 
The curvature formula for a arc-length parameterized space curve is $\kappa ^2= \ddot{x}^2 + \ddot{y}^2$. 
From this it follows that $\kappa^2$ is a $\chi^2$ random variable.

\paragraph{Graph parameterization.}

We also analyze the case where the curve is the graph of a one-dimensional function, i.e., $x=x$,$y=f(x)$. 
In this case the curvature can be written as $\kappa=\ddot{f}/((1+\dot{f}^2)^\frac{3}{2})$. 
Once again all the derivatives $\dot{f}$ and $\ddot{f}$ are Gaussian random variables. 
Assume that $(\dot{f}, \ddot{f})$ are distributed according to $N(0,\Sigma)$. Here $\Sigma = [\sigma_{\dot{f},\dot{f}}, \sigma_{\dot{f}\ddot{f}}; \sigma_{\dot{f}\ddot{f}}, \sigma_{\ddot{f}\ddot{f}} ]$ denoting the joint covariance distribution.
Applying the delta method with $g(a, b) = b/(1+a^2)^{3/2}$, we get that $k$ is distributed as a Gaussian random variable $N(0, \nabla g^T \Sigma \nabla g)$. Since $\nabla g(a, b) |_{0,0} = [0, 1]$, we have $k \rightarrow N(0, \sigma_{\ddot{f}\ddot{f}})$.

{\small
\bibliographystyle{ieee_fullname}
\bibliography{egbib,mrnet}
}

\clearpage
\bibliographystyle{splncs04}
\bibliography{egbib,mrnet,prgan}
\end{document}